\documentclass[10pt, a4paper]{article}
\usepackage{lrec}
\usepackage{graphicx}
\usepackage{tabularx}
\usepackage{soul}
\usepackage{booktabs}
\usepackage{stfloats}
\usepackage{multirow}
\usepackage{CJK}

\usepackage{epstopdf}
\usepackage[latin1]{inputenc}

\usepackage{xcolor}
\definecolor{cR}{RGB}{255,204,204}
\definecolor{cG}{RGB}{204,255,204}
\definecolor{cB}{RGB}{204,204,255}
\definecolor{cK}{RGB}{221,221,221}
\definecolor{navyblue}{rgb}{0.0,0.0,0.5}
\usepackage[hidelinks]{hyperref}
\hypersetup
{
    colorlinks=true,
    citecolor=navyblue,
    linkcolor=navyblue,
    filecolor=navyblue,
    urlcolor=navyblue,
}

\usepackage{xstring}
\makeatletter
\newcommand{\printfnsymbol}[1]{%
  c%
}
\makeatother

\title{JASS: Japanese-specific Sequence to Sequence Pre-training \\ for Neural Machine Translation}

\name{Zhuoyuan Mao$^{\dag}$, Fabien Cromieres$^{\dag}$\textsuperscript{$\star$}\thanks{$\star$ represents equal contribution to this work}, Raj Dabre$^{\ddag}$\textsuperscript{$\star$}, Haiyue Song$^{\dag}$, Sadao Kurohashi$^{\dag}$}

\address{$^{\dag}$Kyoto University\\
         $^{\ddag}$National Institute of Information and Communications Technology\\
         \{zhuoyuanmao, fabien, song, kuro\}@nlp.ist.i.kyoto-u.ac.jp, raj.dabre@nict.go.jp\\}

\abstract{
Neural machine translation (NMT) needs large parallel corpora for state-of-the-art translation quality. Low-resource NMT is typically addressed by transfer learning which leverages large monolingual or parallel corpora for pre-training. Monolingual pre-training approaches such as MASS (MAsked Sequence to Sequence) are extremely effective in boosting NMT quality for languages with small parallel corpora. However, they do not account for linguistic information obtained using syntactic analyzers which is known to be invaluable for several Natural Language Processing (NLP) tasks. To this end, we propose JASS, Japanese-specific Sequence to Sequence, as a novel pre-training alternative to MASS for NMT involving Japanese as the source or target language. JASS is joint BMASS (Bunsetsu MASS) and BRSS (Bunsetsu Reordering Sequence to Sequence) pre-training which focuses on Japanese linguistic units called bunsetsus. In our experiments on ASPEC Japanese--English and News Commentary Japanese--Russian translation we show that JASS can give results that are competitive with if not better than those given by MASS. Furthermore, we show for the first time that joint MASS and JASS pre-training gives results that significantly surpass the individual methods indicating their complementary nature. We will release our code, pre-trained models and bunsetsu annotated data as resources for researchers to use in their own NLP tasks.
\\ \newline  }

\begin{document}
\maketitleabstract
\section{Introduction}
Encoder-decoder based neural machine translation (NMT) \cite{DBLP:journals/corr/SutskeverVL14,DBLP:journals/corr/BahdanauCB14:original}, and in particular, the Transformer model  \cite{NIPS2017_7181} have led to a large jump in the quality of automatic translation over previous approaches such as Statistical Machine Translation \cite{koehn-2004-statistical}. One of the drawbacks of NMT is that it requires large parallel corpora for training robust and high quality translation models. This strongly limits its usefulness for many language pairs and domains for which no such large corpora exist.

The most popular way to solve this issue is to leverage monolingual corpora, which are much easier to obtain (as compared to parallel corpora) for most languages and domains. This can be done either by backtranslation \cite{sennrich-etal-2016-improving,hoang-etal-2018-iterative,edunov-etal-2018-understanding} or  by pre-training. Pre-training consists in initializing some or all of the parameters of the model through tasks that only require monolingual data. One can pre-train the word embeddings of the model \cite{qi-etal-2018-pre} or the encoder and decoders \cite{DBLP:conf/emnlp/ZophYMK16:original}. Pre-training has recently become the focus of much research after the success of methods such as BERT \cite{DBLP:journals/corr/abs-1810-04805}, ELMO \cite{Peters:2018} or GPT \cite{Radford2018ImprovingLU} in many NLP tasks. However, these methods were not designed to be used for NMT models in the sense that BERT-like models are essentially language models and not sequence to sequence models.
 \cite{song2019mass} have obtained new state-of-the-art results for NMT in low-resource settings by addressing these issues and providing a pre-training method for sequence to sequence models: MASS (MAsked Sequence to Sequence).

Another way to overcome the scarcity of parallel data is to provide the model with more ``linguistic knowledge", such as language-specific information. Works such as \cite{sennrich-haddow-2016-linguistic,rudramurthy19,zhou-etal-2019-handling} have shown that such information could improve results. However, because NMT models are end-to-end sequence to sequence models, the manner in which such linguistic hints should be provided is not always clear.

In this paper, we argue that pre-training provides an ideal framework both for leveraging monolingual data and improving NMT models with linguistic information. Our setting focuses on the translation between language pairs involving Japanese. Japanese is a language for which very high quality syntactic analyzers have been developed \cite{kurohashi--EtAl:1994,jumanpp2}. On the other hand, large parallel corpora involving Japanese exist only for a few language pairs and domains. As such it is critical to leverage both monolingual data and the syntactic analyses of Japanese for optimal translation quality.

Our pre-training approach is inspired by MASS, but with more linguistically motivated tasks. In particular, we add syntactic constraints to the sentence-masking process of MASS and dub the resulting task BMASS\footnote{For Bunsetsu-MASS, bunsetsus are one of the elementary syntactic components of Japanese}. We also add a linguistically-motivated reordering task that we dub BRSS (Bunsetsu Reordering Sequence to Sequence). We combine these two tasks to obtain a novel pre-training method tailored for Japanese that we call JASS (Japanese-specific Sequence to Sequence). 

We experiment on the ASPEC Japanese--English  dataset in a variety of settings ranging from 1000 to 1,000,000 parallel sentences. We also experiment with a realistic setting for a difficult language pair, namely, Japanese-Russian. Our results show that JASS by itself is already at least as good as and often better than using the state-of-the-art MASS pre-training. Furthermore, we show that combining MASS and JASS lead to further improvements of up to +1.7 BLEU in low resource settings.

\begin{figure*}
         \centering
         \includegraphics[width=1\textwidth]{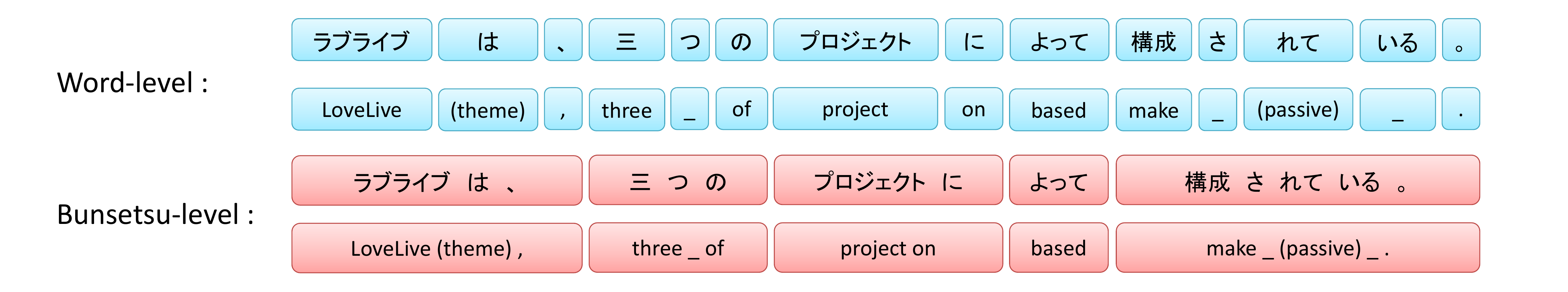}
     \caption{Word and bunsetsu segmentations for a Japanese sentence with the meaning ``LoveLive is made of three projects." Below is the English version translated word-for-word, where ``\_" represents the meaningless segmented parts.}
     \label{fig:bunsetsu}
 \end{figure*}
 
To the best of our knowledge, this is the first time that syntactic information is used in such a pre-training setting for NMT. We make our code and pre-trained models publicly available\footnote{\url{https://github.com/Mao-KU/JASS}}.

The contributions of this paper are as follows:
\begin{itemize}
    \item JASS: a novel linguistically motivated pre-training method for NMT involving Japanese.
    \item Showing how MASS and JASS complement each other indicating that combining multiple types of pre-training techniques can yield better results than using only one type of pre-training.
    \item An empirical comparison of MASS and JASS for ASPEC Japanese--English translation in several data size settings to identify situations where each technique can be most useful.
    \item Verifying that pre-training is a good way to feed linguistic information into to a model. 
    \item Pre-trained models, code and annotated data as resources for reproducibility and public use.
\end{itemize}

\section{Related Work}
Pre-training based approaches are essentially transfer learning approaches where we leverage an external source of data to train a model whose components can be used for NLP tasks which do not have abundant data. In the context of NMT, cross-lingual transfer \cite{DBLP:conf/emnlp/ZophYMK16:original} was shown to be most effective to improve Hausa-English translation when a pre-trained French-English NMT model was fine-tuned on Hausa-English data. While this work focused on strongly pre-training the English side decoder, \cite{dabre-etal-2019-exploiting} showed that pre-training the encoder is also useful through experiments on fine-tuning an English--Chinese model on a small multi-parallel English--X (7 Asian languages) data. All these works rely on bilingual corpora but our focus is on leveraging monolingual corpora that are orders of magnitude larger than bilingual corpora.

Pre-trained models such as BERT \cite{DBLP:journals/corr/abs-1810-04805}, ELMO \cite{Peters:2018}, XLNET \cite{DBLP:journals/corr/abs-1906-08237} and GPT \cite{Radford2018ImprovingLU}  have proved very useful for tasks such as Text Understanding, but have a limited application to NMT, as they only pre-train the encoder side of a transformer. Pre-training schemes more suitable to NMT have been proposed by \cite{DBLP:journals/corr/abs-1901-07291},  \cite{ren-etal-2019-explicit} and \cite{song2019mass}. In particular, \cite{song2019mass} obtained state-of-the-art results with their ``MASS" pre-training scheme. MASS allows for the simultaneous pre-training of the encoder and decoder and hence is the most useful for NMT. However, MASS does not consider the linguistic properties of language when pre-training whereas our objective is to show that linguistically motivated pre-training can be complementary to MASS. Our research is motivated by previous research~\cite{kawahara-etal-2017-automatically} for Japanese NLP which showed that linguistic annotations from the syntactic analyzers such as Juman \cite{jumanpp2} and KNP \cite{kurohashi--EtAl:1994} are extremely important.

Pre-ordering consists of pre-processing a sentence so that its word-order is more similar to that of its expected translation. It has been a popular technique for Statistical Machine Translation since the early work of \cite{collins2005clause}. Although initial research \cite{du2017pre} had concluded that pre-ordering had limited usefulness for NMT,  it has been shown more recently that it can improve translation quality, especially in the case of low-resource languages. \cite{rudramurthy19} showed that pre-ordering English to Indic language word order is beneficial when performing transfer learning via fine-tuning. \cite{zhou-etal-2019-handling} showed that leveraging structural knowledge for creating the psuedo Japanese-ordered English by pre-ordering English from SVO to SOV improves Japanese--English translation. Our work will try to incorporate similar ideas directly in the pre-training process. On the related matter of the usefulness of linguistic information for NMT, \cite{sennrich-haddow-2016-linguistic} also showed how linguistic annotations can help improve German--English translation. 

\section{Background: MASS and Bunsetsu}
Central to our work are Bunsetsu and MASS which we explain as below.

\begin{figure*}
\begin{center}
\includegraphics[width=1\linewidth]{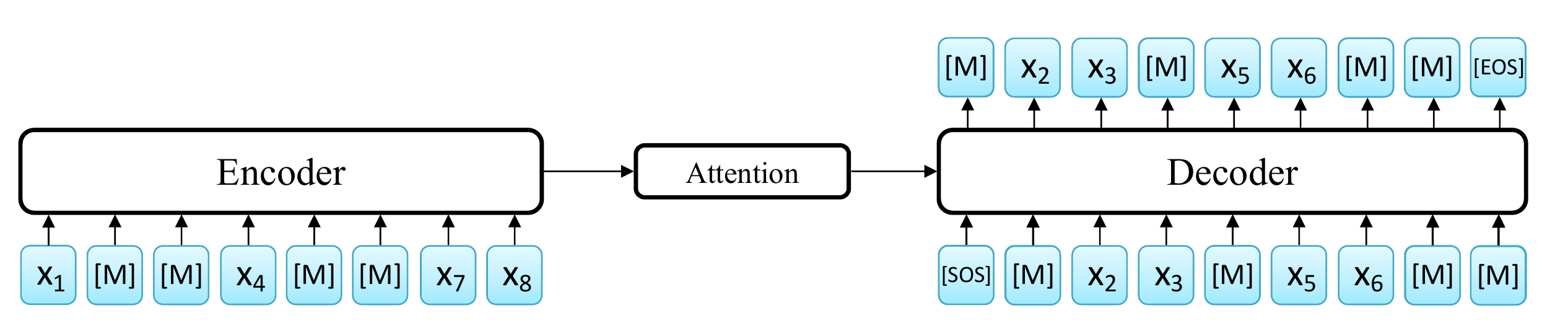} 
\caption{Sequence to Sequence Structure for MASS, where $x_i$ represents a token and $x_2, x_3$ and $x_5, x_6$ are consecutive tokens to be masked/predicted. In the case of BMASS pre-training, $x_2, x_3$ and $x_5, x_6$ are bunsetsus.}
\label{fig.1}
\end{center}
\end{figure*}

\subsection{Bunsetsu}
Bunsetsus are syntactic components of Japanese sentences. They are roughly equivalent to the noun chunks or verb chunks in English syntax. They constitute a minimal unit of meaning.  Japanese segmenters can segment a Japenese sentence in words or in bunsetsus, but the concept of ``word" is ambiguous for writing systems that do not use word-separators (like spaces), like Japanese. Bunsetsus are also more likely to correspond to a well-defined entity or concept than words. The conceptual difference between using word level and bunsetsu level segmentation is shown in Figure~\ref{fig:bunsetsu} for the Japanese sentence with the meaning ``LoveLive is made of three projects." 
Note that each bunsetsu contains some self contained information and some case marker which can indicate its relation with another bunsetsu.

\subsection{MASS}
MASS is a pre-training method for NMT proposed by \cite{song2019mass}.
In MASS pre-training the input is a sequence of tokens where a part of the sequence is masked and the output is a sequence where the masking is inverted. Consider $x\in \mathcal{X}$ which is a sequence of tokens where $\mathcal{X}$ is a monolingual corpus. Consider $C = [[p_{1},p_{2}],[p_{3},p_{4}],...[p_{n},p_{n+1}]]$ where $0< p_{1}\leq p_{2}\leq p_{3}\leq p_{4}\leq ... p_{n}\leq p_{n+1}\leq len(x)$ and $len(x)$ is the number of tokens in sentence $x$.
We denote by $x^{C}$ the masked sequence where tokens in positions from $p_{1}$ to $p_{2}$, $p_{3}$ to $p_{4}$ and so on until $p_{n}$ to $p_{n+1}$ in $x$ are replaced by a special token $[M]$. $x^{!C}$ is the invert masked sequence where tokens in positions other than the aforementioned fragments are replaced by the mask token $[M]$. 
MASS is a pre-training objective that predicts the masked fragments in $x$ using an encoder-decoder model where $x^{C}$ is the input to the encoder and $x^{!C}$ is the reference for the decoder. The log likelihood objective function is:


\begin{eqnarray}
\label{eqn:mass}
\mathcal{L}_{mass}(\mathcal{X}) &=& \frac{1}{|\mathcal{X}|} \sum_{x \in \mathcal{X}} \log P\left(x^{!C} | x^{C}; \theta\right)  \\
&=& \frac{1}{|\mathcal{X}|} \sum_{x \in \mathcal{X}} \log \prod_{t \in C} P\left(x_{t}^{!C} | x_{<t}^{!C}, x^{C} ; \theta\right)  \nonumber
\end{eqnarray}

where $x_{<t}^C$ indicates the preceding tokens before $t$ in $x^C$ and $\theta$ is set of model parameters.
The hyper-parameter for MASS is the number of tokens to be masked. Refer to Figure~\ref{fig.2}-b for a training pair example for MASS.


\section{Proposed Method: JASS}
JASS (Japanese-specific Sequence to Sequence pre-training) is an extension of the original MASS method to incorporate linguistic information in addition to reordering based pre-training \cite{zhang-zong-2016-exploiting}. It is a combination of two sub-methods, BMASS (Bunsetsu-based MAsked Sequence to Sequence pre-training) and BRSS (Bunsetsu Reordering Sequence to Sequence pre-training). 

\subsection{BMASS} \label{sec:bmass}
In MASS, a NMT model is trained by making it predict random parts of a sentence given their context. Instead of random parts we are interested in making the model predict a set of bunsetsus given the contextual bunsetsus. We expect this will let the model learn about the important concept of bunsetsu, as well as focus its training on predicting meaningful subsequences instead of random ones.

More precisely, we propose BMASS (Bunsetsu-based MAsked Sequence to Sequence pre-training), which leverages syntactic parses of Japanese monolingual data for sequence to sequence pre-training. To perform BMASS, we modify the mask $C$ in Equation~\ref{eqn:mass} where the position spans $p_{1}$ to $p_{2}$, $p_{3}$ to $p_{4}$ and so on until $p_{n}$ to $p_{n+1}$ indicate the start and end of bunsetsus in a Japanese sentence. Consequently we denote the BMASS loss as $\mathcal{L}_{bmass}$. The main difference between MASS and BMASS is that in MASS we mask random token spans whereas in BMASS we mask tokens spans that only cover bunsetsus. The number of bunsetsus to be masked constitutes a hyper-parameter for BMASS.

\begin{figure*}
\begin{center}
\includegraphics[width=1\linewidth]{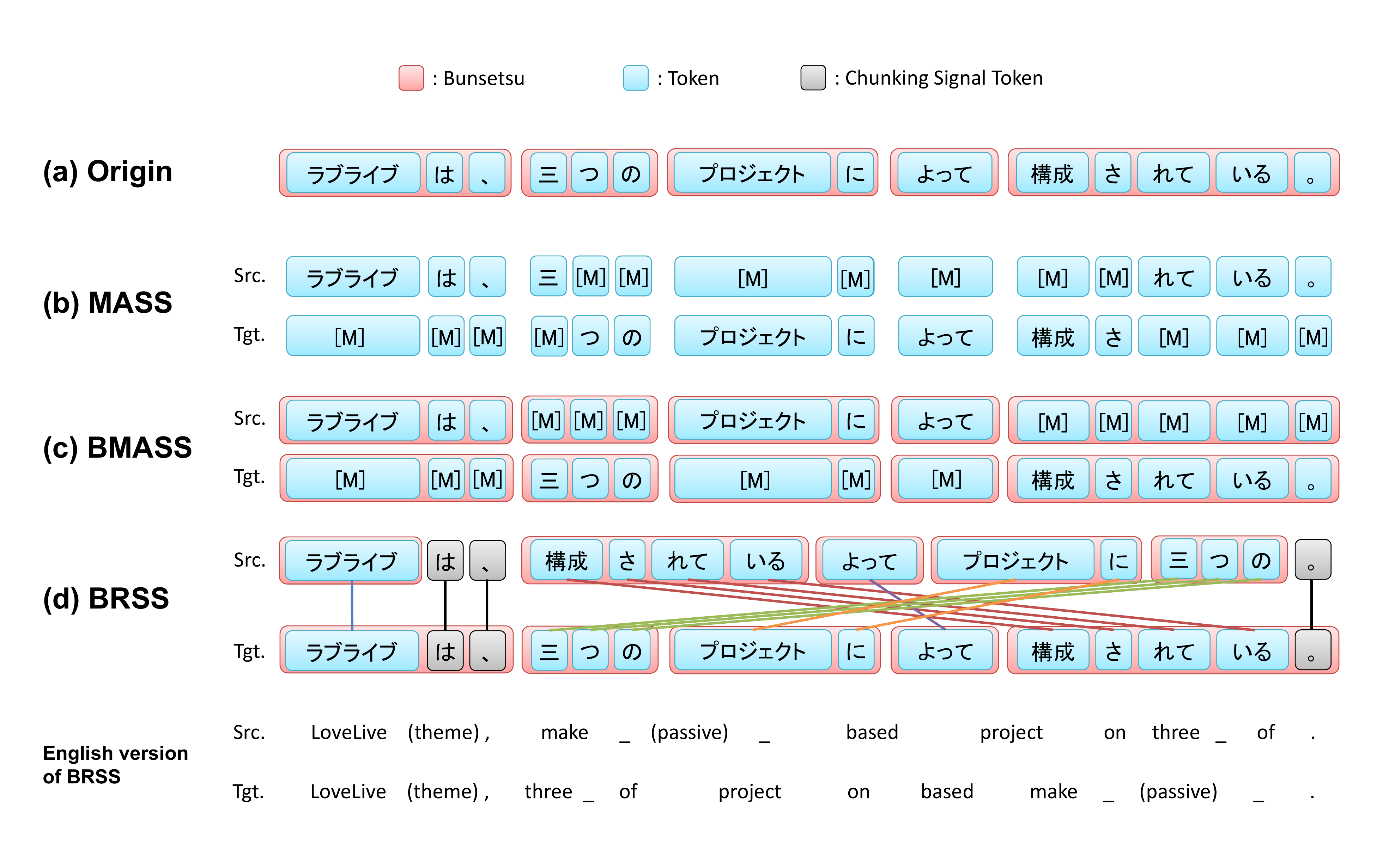} 
\caption{An example of source and target for MASS, BMASS, BRSS with the meaning ``LoveLive is made of three projects."}
\label{fig.2}
\end{center}
\end{figure*}

Refer to Figure~\ref{fig.2}-c for a training pair example for BMASS, which may be contrasted with the MASS example in figure~\ref{fig.2}-b.

\subsection{BRSS} \label{sec:rss}
BRSS (Bunsetsu Reordering based Sequence to Sequence) roughly consist in training the NMT system with re-ordered Japanese. We expect that this will let the system learn the structure of  Japanese language, as well as prepare it for the reordering operation it will have to perform when translating to a language with different grammar.

\subsubsection{Bunsetsu-based Reordering}\label{sec:breordering}
We first define here a simple process for re-ordering a (typically SOV) Japanese sentence into a ``SVO-ordered Japanese" pseudo-sentence. We will then use this reordered sentence in section~\ref{sec:brsspretraining} for our BRSS pre-training.

There exist several previous works about reordering a SOV-ordered sentence to a SVO-ordered sentence \cite{DBLP:conf/ntcir/Katz-BrownC08,hoshino-etal-2014-japanese}. In our case, in order to leverage bunsetsu units in Japanese consistently with BMASS, we propose Bunsetsu-based Reordering, which is able to generate a SVO-ordered Japanese sentence while retaining syntactic information at the bunsetsu-level. Bunsetsu-based Reordering is implemented by the following steps: 
\begin{itemize}
\item Split the Japanese into several chunks by chunking signal tokens. Specifically, chunking signal tokens includes the punctuation and `\begin{CJK}{UTF8}{min}は\end{CJK}' which mentions the theme in a Japanese sentence
\item Reverse the order  of the bunsetsus in each chunk
\end{itemize}


\subsubsection{Bunsetsu Reordering Sequence to Sequence Pre-training}\label{sec:brsspretraining}
We build up our BRSS (Bunsetsu Reordering Sequence to Sequence) pre-training on the basis of Bunsetsu-based Reordering as mentioned above. Refer to Figure~\ref{fig.2}-d for a training pair example for BRSS. The pre-training objective here is a deshuffling or un-reordering task which reconstructs the original sentence from the reordered sentence.  

The bunsetsu reordering process described above in section~\ref{sec:breordering} allows us to produce an artificial ``SVO-Japanese" sentence for each sentence in a training monolingual corpora. We then have two choices for the pre-training procedure. We can make the NMT system predict the artificial reordered  SVO sentence given the original. Alternatively, we can make it predict the original given the reordered one.


We experiment with both options in the following section. These 2 pre-training directions are denoted as BRSS.F (reordered to original) and BRSS.R (original to reordered). Following the notation in MASS, we define the log likelihood objective function of BRSS as follows:
\begin{eqnarray}
\mathcal{L}_{brss}(\mathcal{X}) &=& \mathcal{L}_{brss.f}(\mathcal{X}) \;   or \;   \mathcal{L}_{brss.r}(\mathcal{X}) \\
\mathcal{L}_{brss.f}(\mathcal{X}) &=& \frac{1}{|\mathcal{X}|} \sum_{x \in \mathcal{X}} \log P\left(x | x^{reordered}; \theta\right) \\ 
\mathcal{L}_{brss.r}(\mathcal{X}) &=& \frac{1}{|\mathcal{X}|} \sum_{x \in \mathcal{X}} \log P\left(x^{reordered} | x; \theta\right) 
\end{eqnarray}

Note that the equivalent reordering based pre-training mechanism is one where we randomly shuffle a sentence at the word level. However, this does not focus on learning any kind of specific linguistic reordering and so we do not explore it in this work.


\subsection{JASS}
In the previous sections, we have defined two pre-training procedures: BMASS and BRSS. Our actual pre-training will consist in a joint execution of these two pre-training. We call the resulting pre-training JASS (JApanese-specific Sequence to Sequence) pre-training. The pre-training objective for JASS is therefore:
\begin{eqnarray}
\mathcal{L}_{jass}(\mathcal{X}) = \mathcal{L}_{bmass}(\mathcal{X}) + \mathcal{L}_{brss}(\mathcal{X})
\end{eqnarray}
where $\mathcal{X}$ represents the monolingual corpus of Japanese. We expect BMASS to learn syntactic knowledge and BRSS to learn word ordering knowledge. 

Because JASS has been specifically designed for Japanese, and we have not yet considered equivalents for other languages, we also mix JASS pre-training for Japanese with MASS pre-training for the other language involved in the translation.

In practice, we therefore designate by JASS the pre-training of the NMT system that uses Japanese monolingual data with BMASS and BRSS objectives, and ``other language" monolingual data with MASS objective.

We can also consider using Japanese monolingual data with a combination of BMASS, BRSS and MASS objectives, which we dub MASS+JASS in the following sections.

\section{Experimental Settings}
In this section, we evaluate our pre-training methods on 4 translation directions: Japanese-to-English (Ja-En), English-to-Japanese (En-Ja), Japanese-to-Russian (Ja-Ru) and Russian-to-Japanese (Ru-Ja). Specifically, we monitor the performance of our pre-training methods on both simulated low-resource and high-resource scenarios involving ASPEC Japanese--English translation \cite{nakazawa-EtAl:2015:WAT}. We also test our methods on a realistic low-resource scenario involving News Commentary Japanese--Russian translation\footnote{Neither Japanese nor Russian are low-resource languages, but Ja-Ru can be regarded as a low-resource language pair because of the limited amount of the parallel data.} \cite{imankulova-etal-2019-exploiting}.

\subsection{Datasets and Pre-processing}
We use both the monolingual data and parallel data for pre-training and the parallel data for fine-tuning. Refer to Table~\ref{data} for an overview. \begin{table}
\begin{center}
\begin{tabular}{c|c |c| c}

      \toprule
      & Language& Dataset&Size\\
       \midrule
      \multirow{3}{*}{Mono}&Ja& Common Crawl& 22M\\
      &En& News Crawl& 22M\\
      &Ru& News Crawl& 22M\\
       \midrule
      \multirow{2}{*}{Parallel}&Ja-En& ASPEC-JE&3M\\
     &Ja-Ru&JaRuNC&10K\\
      \bottomrule
\end{tabular}
\caption{Overview of data} 
\label{data}
 \end{center}
\end{table}

\subsubsection{Parallel Data}
We use scientific abstracts domain ASPEC parallel corpus \cite{NAKAZAWA16.621} for Japanese--English translation and the news commentary domain JaRuNC parallel corpus \cite{imankulova-etal-2019-exploiting} for Japanese--Russian translation.

\subsubsection{Monolingual data}
We use monolingual data containing 22M Japanese, 22M English and 22M Russian sentences randomly sub-sampled from Common Crawl dataset and News crawl\footnote{The pre-training will be very effective if the domains of  the pre-training and fine-tuning dataset are similar\cite{raffel2019exploring}. However, in order to obtain a general pre-trained model for NMT, we choose the monolingual data from Common Crawl and News Crawl.} dataset from the official WMT monolingual training data\footnote{\url{http://www.statmt.org/wmt19/translation-task.html}} for pre-training. Each side of the parallel data used in fine-tuning is also incorporated into the monolingual data for pre-training. Specifically, for Japanese and English, 3M sentences from each side of the parallel data is added to the monolingual data while for Japanese and Russian, 10K sentences from each side of the parallel data is also used in pre-training. This results in 50M monolingual sentences for Japanese and English, and 45M monolingual sentences for Japanese and Russian. Given that our pre-training objective works at the monolingual level and that the three languages have different scripts and thus have few common words, we believe this to be a fair pre-training data setting.

\subsubsection{Pre-processing}
We tokenize the monolingual data by using the Moses tokenizer\footnote{\url{https://github.com/moses-smt/mosesdecoder}} for En and Ru, and the Jumanpp tokenizer\footnote{\url{https://github.com/ku-nlp/jumanpp}} for Ja. We get the bunsetsu information by using KNP\footnote{\url{https://github.com/ku-nlp/pyknp}}. 
Sentences with length over 175 tokens are removed.
For each language pair, we built a joint vocabulary with 60,000 sub-word units via Byte-Pair Encoding\cite{sennrich-etal-2016-neural}. Considering the discrepancy of the domain between pre-training dataset and fine-tuning dataset, we oversample the fine-tuning dataset when learning BPE codes. Since some English alphabets appear in the Japanese and Russian corpora, the BPE codes are learned jointly from the concatenation of the corpora for each language pair. As we do multi-task pre-training, each sentence is prepended with a task token $[MASS]$, $[BMASS]$ or $[RSS]$ and a language token $[Ja]$, $[En]$, or $[Ru]$. This ensures that the model learns to distinguish between different pre-training objectives and languages.

\subsection{Model Training and Evaluation Settings}
For the NMT model, we experiment with a Transformer \cite{NIPS2017_7181}  having 6 layers for both the encoder and the decoder. We implement our approaches on top of the OpenNMT\footnote{\url{https://github.com/OpenNMT/OpenNMT-py}} transformer implementation. 

\begin{table}[h!]
\begin{center}
\begin{tabularx}{\columnwidth}{c| c c c c c}

      \toprule
      Model & 1K & 10K & 20K & 100K & 1M\\
      \midrule
      Transformer-big & 0.40 & 2.56 & 9.53 & 22.72 & 29.50\\
      Transformer-base & 0.33 & 1.79 & 8.21 & 21.34 & 29.06\\
      \bottomrule
\end{tabularx}
\caption{BLEU on Ja-En (ASPEC-JE)} 
\label{big}
 \end{center}
\end{table}

OpenNMT provides two default hyperparameters settings that  differ in the size of layer used and the number of attention heads, namely, ``base" and ``big". Although we could have expected the smaller model to be a better fit for low-resource training, we found out the opposite. Table~\ref{big} contains our preliminary experiments where the Transformer in the big setting outperforms Transformer in the base setting for both high-resource and low-resource scenarios for Japanese--English translation. 

Therefore, we implement our pre-training methods and fine-tuning using the Transformer-big setting, which consists of a 6-layer encoder and a 6-layer decoder, with the length of 1024 for hidden size, the length of 4096 for feed-forward size, dropout rate of 0.3 and attention heads of 16. A learning-rate of $10^{-4}$ is used both for pre-training and fine-tuning, and all the pre-training tasks are implemented on 8 TITAN X (Pascal) GPU cards until convergence with a batch-size of 2048 for each GPU while single GPU is used for fine-tuning. The checkpoint with the highest accuracy is selected for fine-tuning. We use BLEU \cite{papineni-etal-2002-bleu} to implement the evaluation. We do early stopping if no improvement on development-set within 5 checkpoints, and the checkpoint with the best BLEU performance on development-set is selected for evaluation.\\
For multi-task pre-training, data is randomly shuffled so that even in each mini-batch, different pre-training objectives will appear, corresponding to a real joint pre-training.

 We evaluate the statistical significance of our BLEU scores by bootstrap resampling \cite{koehn-2004-statistical}.

\subsection{Pre-trained models}
We pre-train our NMT models by leveraging the monolingual data of the source and target languages. For Japanese we use MASS as well as JASS, while for English and Russian, we only use MASS as the pre-training objective. In particular we pre-train the following models:

\begin{itemize}
\item \textbf{MASS:} We use the same settings as in \cite{song2019mass} for pre-training.
\item \textbf{BMASS:} Similar to MASS we mask half the bunsetsus in a sentence during pre-training.
\item \textbf{BRSS:} Using our approach in Section~\ref{sec:rss} we pre-train on SVO--SOV (BRSS.F) Japanese sentence pairs.
\item \textbf{JASS:} Multi-task training of BMASS and BRSS.
\item \textbf{MASS+BMASS:} Multi-task training of MASS and BMASS.
\item \textbf{MASS+BRSS:} Multi-task training of MASS and BRSS.
\item \textbf{JASS+MASS:} Multi-task training of BMASS, BRSS and MASS.
\end{itemize}


\subsection{Fine-tuning on NMT}
As mentioned above, we validate the effectiveness of our pre-training methods by 4 fine-tuning tasks, which are Ja-En, En-Ja, Ja-Ru, Ru-Ja. We train the following models by fine-tuning the pre-trained models:
\begin{itemize}
    \item \textbf{ASPEC Ja--En and En--Ja:} Japanese to English and English to Japanese models using from 1K to 1M\footnote{We limit ourselves to 1M sentences because the remaining 2M sentences are relatively noisy and most of previous research mainly relies on the best 1M sentences for best translation quality.} parallel sentences.
    \item \textbf{NC Ja--Ru and Ru--Ja:} Japanese to Russian and Russian to Japanese models using available 12,356 training pairs .
\end{itemize}
We compare these models with baselines which do not use pre-training.

\section{Results \& Analysis}

We now give the results for Japanese--English and Japanese--Russian translation. All the results are reported on the official test sets provided by the 2019 edition of the Workshop on Asian Translation(WAT)\footnote{\url{http://lotus.kuee.kyoto-u.ac.jp/WAT/WAT2019/index.html}}.

\subsection{Pre-training Accuracy}
\begin{table}[h!]
\begin{center}
\begin{tabular}{c |c |c }

      \toprule
      Setting & En+Ja & Ru+Ja\\
      \midrule
      MASS &71.18&72.35\\
      \midrule
      BMASS&73.76&73.98\\
      BRSS.F&84.82&84.89\\
      JASS(BMASS+BRSS.F)&81.53&81.63\\
      \midrule
      MASS+BMASS &72.33&-\\
      MASS+BRSS.F&79.56&-\\
      MASS+JASS&78.62&78.85\\
      \bottomrule
\end{tabular}
\caption{Pre-training accuracy, which is the 1-gram accuracy of the pre-trained model} 
\label{acc}
 \end{center}
\end{table}
Pre-training accuracy can be an indicator of the learning difficulty. The pre-training objectives should not be too easy or too difficult. As shown in Table~\ref{acc}, for Japanese--English pre-training, BRSS is the easiest for the neural network while MASS is of the highest difficulty. Moreover, it can be found that the accuracy of a pre-training objective does not vary a lot from one language pair to another. As easy and difficult are subjective we use pre-training accuracy as one of the indicators of the difficulty and hence the usefulness of our pre-training approach. MASS+JASS gives the best BLEU performance in most of our experiments and thus we hypothesize that there is no perfect pre-training method and thus one should explore a variety of methods for a given language pair.

\subsection{Fine-tuning Results}

\begin{table*}[!ht]
\begin{center}
\begin{tabular}{c | c c c c c c| c c c c}

      \toprule
      Model & 1K & 3K & 6K & 10K & 20K & 50K & 100K & 200K & 500K & 1M\\
      \midrule
      Supervised(Transformer-big) & 0.40&	1.30&1.55&	2.56	&9.53&	17.56	&22.72&	25.51&	27.92&	29.50\\
      MASS &5.34	&9.89&	12.28&	15.16&	18.65&	22.28&	24.86	&26.67&	28.85&	29.63\\
      \midrule
      BMASS&4.06&	8.49&	11.70&	14.32&	18.56&	22.30&	24.65&	26.77&	28.55&	29.72\\
      BRSS.F&3.29&	8.61&	12.12&	14.75&	18.40&	22.07&	24.55&	26.53&	28.71&	29.53\\
      BRSS.R &3.00	&7.36&	11.02	&13.74&	17.30&	21.80&	24.52	&26.56&	28.45&	29.52\\
      JASS(BMASS+BRSS.F)&5.18&	10.06&	13.49$^\dag$&	15.55$^\dag$&	19.12$^\dag$&	22.85$^\dag$&	\textbf{25.20}&	26.88&	28.62&	29.67\\
      \midrule
      MASS+BMASS&4.64&	8.74&	12.39&	14.22&	18.18&	22.21&	24.86&	26.68&	28.96&29.80\\
      MASS+BRSS.F&5.88$^{\dag}$&	\textbf{10.78}$^{\dag}$&	13.53$^{\dag}$&	15.99$^{\dag}$&	19.01&	22.67&	24.90&	26.75&	\textbf{28.98}&\textbf{29.88}\\
      MASS+JASS&\textbf{6.28}$^\dag$&	10.72$^\dag$&	\textbf{13.97}$^\dag$&	\textbf{16.09}$^\dag$&	\textbf{19.34}$^\dag$&	\textbf{23.15}$^\dag$&	24.99&	\textbf{27.09}$^\dag$&	28.82&29.49\\
      \bottomrule
\end{tabular}
\caption{BLEU scores for simulated low/high-resource settings for Ja-En ASPEC translation using
3K to 1M parallel sentences for fine-tuning. Results better than MASS with statistical significance $p <0.05$ are marked with \dag} 
\label{ja-en}
 \end{center}
\end{table*}

\begin{table*}[!ht]
\begin{center}
\begin{tabular}{c |c c c c c c |c c c c}

      \toprule
      Model & 1K & 3K & 6K & 10K & 20K & 50K & 100K & 200K & 500K & 1M\\
      \midrule
      Supervised(Transformer-big) &0.75&	1.49&	2.21&	3.68&	11.52&	20.95&	27.94&	32.71&	38.89&	40.26\\
      MASS &5.81&	11.02&	15.29&	18.11&	21.57&	27.91&	31.62&	34.88&	38.97&	\textbf{41.16}\\
      \midrule
      BMASS&5.03&	9.77&	13.40&	17.25&	21.14&	27.10&	30.97&	34.90&	39.00&	40.50\\
      BRSS.F&3.54&	10.30&	14.86&	17.67&	21.64&	27.48&	31.22&	34.88	&38.21&	40.43\\
      BRSS.R &4.31	&9.77&	14.25&	16.89&	20.81&	26.34	&30.69&	33.91&	38.49&	40.27\\
      JASS(BMASS+BRSS.F)&5.54&	11.37&	15.91$^\dag$&	18.50$^\dag$	&22.18$^\dag$&	27.27&	31.05&	34.72	&38.89&	40.64\\
      \midrule
      MASS+BMASS&5.20	&10.00&	14.37	&17.44&	21.53&	27.24&	30.98&	\textbf{35.14}&	\textbf{39.40}$^{\dag}$&40.65\\
      MASS+BRSS.F&6.53$^{\dag}$	&12.04$^{\dag}$&	15.79$^{\dag}$	&18.95$^{\dag}$&	22.32$^{\dag}$&	27.32&	\textbf{31.63}&	34.69&	38.85&41.09\\
      MASS+JASS&\textbf{6.82}$^\dag$&	\textbf{12.57}$^\dag$&	\textbf{16.22}$^\dag$&	\textbf{19.20}$^\dag$&	\textbf{23.00}$^\dag$&	\textbf{28.09}&	31.43&	34.81&	38.43&40.79\\
      \bottomrule
\end{tabular}
\caption{BLEU scores for simulated low/high-resource settings for Ja-En ASPEC translation using
3K to 1M parallel sentences for fine-tuning. Results better than MASS with statistical significance $p <0.05$ are marked with \dag} 
\label{en-ja}
 \end{center}
\end{table*}

\begin{table}[]
\begin{center}
\begin{tabular}{c |c| c}

      \toprule
      Model & Ja-Ru & Ru-Ja\\
      \midrule
      Supervised(Transformer-big) &0.50	&0.72\\
      MASS &0.96&	2.84\\
      \midrule
      BMASS&0.97&	2.77\\
      BRSS.F &0.85&	2.36\\
      JASS(BMASS+BRSS.F) &\textbf{1.20	}&3.08\\
      \midrule
      MASS+JASS&1.07&\textbf{3.45}$^{\dag}$\\
      \bottomrule
\end{tabular}
\caption{BLEU scores for Ja-Ru translation on JaRuNC.  Results better than MASS with statistical significance $p <0.05$ are marked with \dag} 
\label{jaru}
 \end{center}
\end{table}
Tables~\ref{ja-en},~\ref{en-ja},~\ref{jaru} contain the results of fine-tuning the pre-trained models for Japanese--English and English--Japanese translation. Our pre-training methods, BMASS and BRSS, clearly improved on the strictly-supervised baselines and fine-tuning gives results comparable to those of MASS, which validate the effectiveness of our Japanese-specific objectives for pre-training. In Table~\ref{ja-en},~\ref{en-ja}, we observe that JASS significantly outperforms ($p < 0.05$)  MASS, when parallel corpora sizes from 3K to 50K are used. In other size settings, JASS is competitive with if not significantly better than MASS and this demonstrates that linguistically motivated pre-training can be an alternative to language-agnostic pre-training. In Table~\ref{ja-en},~\ref{en-ja},~\ref{jaru}, the joint pre-training of MASS and JASS (BMASS+BRSS) leads to the highest BLEU scores ($p < 0.05$) on most settings, which indicates that JASS is not just a alternative to MASS, but could be complementary to MASS. Human evaluation of the translations from both systems should shed more light on this. We leave this for future work. 

Finally, it can be seen that pre-training outperforms the supervised baselines in almost all data scenarios which shows that for neural machine translation, pre-training is a valuable strategy especially for low-resource scenarios. Moreover, since pre-training enables the encoder-decoder to learn an implicit language model it can help overcome the scarcity of language modeling information in parallel corpora. Given the success of JASS in low-resource scenarios, we believe that it is absolutely necessary to leverage language specific information during pre-training. 

Unfortunately, in Table~\ref{jaru}, the improvement contributed by bilingual pre-training is limited on JaRuNC. Japanese--Russian is a difficult language pair, the fine-tuning data is small and the news commentary domain is much harder than the ASPEC domain. As such we feel that multi-lingual pre-training and fine-tuning mechanisms might help alleviate this issue as shown by \cite{imankulova-etal-2019-exploiting}. This too, we leave for future work.

\subsubsection{BRSS.F or BRSS.R?}

Although we mentioned 2 pre-training methods involving reordering which are named as BRSS.F and BRSS.R, we mostly experimented with joint pre-training using BRSS.F. In order to demonstrate this choice, we give the fine-tuning results of BRSS.F and BRSS.R on ASPEC as shown in Table~\ref{ja-en},~\ref{en-ja}. We observe that BRSS.F outperforms BRSS.R in most cases, regardless of translation direction, which is probably because BRSS.F is able to pre-train a better decoder to generate natural language, even though the reordering described by BRSS.R should be more appropriate for Ja-En translation. Currently, we do not have any detailed explanation why BRSS.F is consistently better than BRSS.R and we will investigate this in the future.

\section{Conclusion}
In this paper we proposed JASS (Japanese-specific sequence to sequence) pre-training which are novel pre-training alternatives to MASS for neural machine translation involving Japanese as the source or target language. Our work is aimed at leveraging abundant monolingual data and syntactic analyses provided by analyzers so that the pre-training stage becomes aware of language structure. 
Our experiments on ASPEC Japanese--English translation and News Commentary Japanese--Russian translation have shown that JASS, which leverages syntactic parsing knowledge from the KNP parser, outperform MASS, which is language agnostic, in many low-resource settings. Furthermore, we show that the combination of MASS and JASS yields significantly better results than the individual pre-training methods. This demonstrates the effectiveness of our methods and the necessity to inject language-specific information into the pre-training objective. We have publicly released our code and models. To the best of our knowledge, this is the first time that linguistic information has been used for pre-training a NMT system. Our positive results show that the pre-training step is an appropriate place to provide linguistic hints to a NMT system.

We are now working on several directions for improving and broadening our approach. Pre-training methods do not often consider domain differences within the data and so in the future, we will try to address domain adaptation in order to enhance the impact of fine-tuning on in-domain data. In particular we find the multi-stage training approach \cite{imankulova-etal-2019-exploiting,dabre-etal-2019-exploiting} most relevant in this direction. We will also work on determining the impact of multi-task pre-training using a combination of a wide variety of pre-training approaches that focus on different aspects of language structure. We might also apply ideas similar as the ones developped here to different languages. We also note that \cite{raffel2019exploring} has recently shown that many NLP tasks such as Text Understanding could be reformulated as Text-to-Text tasks. This broadens a lot the domain of usefulness of text-to-text pre-training tasks such as ours, and we will be interested in evaluating our approach on a wider range of NLP tasks.


\section{References}
\bibliographystyle{lrec}
\bibliography{lrec2020W-xample}


\end{document}